\documentclass[letterpaper, 10 pt, conference]{ieeeconf}  

\IEEEoverridecommandlockouts                              

\usepackage{amsfonts}
\usepackage{graphicx}
\usepackage{textcomp} 
\usepackage{multirow}
\usepackage{amsmath}

\title{\LARGE \bf
	A Vision-based Irregular Obstacle Avoidance Framework via Deep Reinforcement Learning
}

\author{Lingping Gao\textsuperscript{1,$*$}, Jianchuan Ding\textsuperscript{1,$*$}, Wenxi Liu\textsuperscript{2}, Haiyin Piao\textsuperscript{3}, Yuxin Wang\textsuperscript{1}, Xin Yang\textsuperscript{1,$\dagger$}, Baocai Yin\textsuperscript{1}
\thanks{\textsuperscript{$*$} Joint first authors. \textsuperscript{$\dagger$} Corresponding author.}
\thanks{This work was supported in part by the National Natural Science Foundation of China under Grant 91748104, Grant 61972067, and the Innovation Technology Funding of Dalian (Project No.2018J11CY010, 2020JJ26GX036).}
\thanks{\textsuperscript{1} School of Computer Science, Dalian University of Technology, Dalian 116024, China  \texttt{\{gaolingping,djc\_dlut\}@mail.dlut.edu.cn}; \texttt{\{wyx,xinyang,ybc\}@dlut.edu.cn}}
\thanks{\textsuperscript{2} College of Mathematics and Computer Science, Fuzhou University, Fuzhou 350108, China   \texttt{wenxi.liu@hotmail.com}}
\thanks{\textsuperscript{3} School of Electronics and Information, Northwestern Polytechnical University, Xi'an 710129, China \texttt{haiyinpiao@mail.nwpu.edu.cn}}}

\begin{document}

	\maketitle
	\thispagestyle{empty}
	\pagestyle{empty}

	\begin{abstract}
	
	 Deep reinforcement learning has achieved great success in laser-based collision avoidance work because the laser can sense accurate depth information without too much redundant data, which can maintain the robustness of the algorithm when it is migrated from the simulation environment to the real world. However, high-cost laser devices are not only difficult to apply on a large scale but also have poor robustness to irregular objects, \emph{e.g.}, tables, chairs, shelves, \emph{etc}. In this paper, we propose a vision-based collision avoidance framework to solve the challenging problem. Our method attempts to estimate the depth and incorporate the semantic information from RGB data to obtain a new form of data, pseudo-laser data, which combines the advantages of visual information and laser information. Compared to traditional laser data that only contains the one-dimensional distance information captured at a certain height, our proposed pseudo-laser data encodes the depth information and semantic information within the image, which makes our method more effective for irregular obstacles. Besides, we adaptively add noise to the laser data during the training stage to increase the robustness of our model in the real world, due to the estimated depth information is not accurate. Experimental results show that our framework achieves state-of-the-art performance in several unseen virtual and real-world scenarios.

	\end{abstract}


	\vspace{-0.4cm}
	\section{Introduction}
	\vspace{-0.1cm}
Collision avoidance is a challenging research problem for robot navigation, which has been studied for decades and can be applied in many important applications in real-world scenarios.
In recent years, deep reinforcement learning (DRL) based methods~\cite{tai2016deep, choi2019deep} have been studied to address the collision avoidance problem in the robot system.
Compared with deep learning methods~\cite{pfeiffer2017perception, chen2017socially}, DRL-based methods do not rely on manually-labeled data. Specifically, in virtual environments, the agent can continuously interact with the environment in a `trial-error' manner and obtain environmental feedback for learning. As the major limitation of learning policy using simulation data, the large gap between simulation and the real-world makes it difficult to directly migrate the learned policy from virtual agents to real robots. One feasible solution is to apply lasers to sense the surrounding environment and formulate the observation as one-dimensional distance information.

\begin{figure}[t]
	\centering\includegraphics [height=49.95mm,width=75mm ]{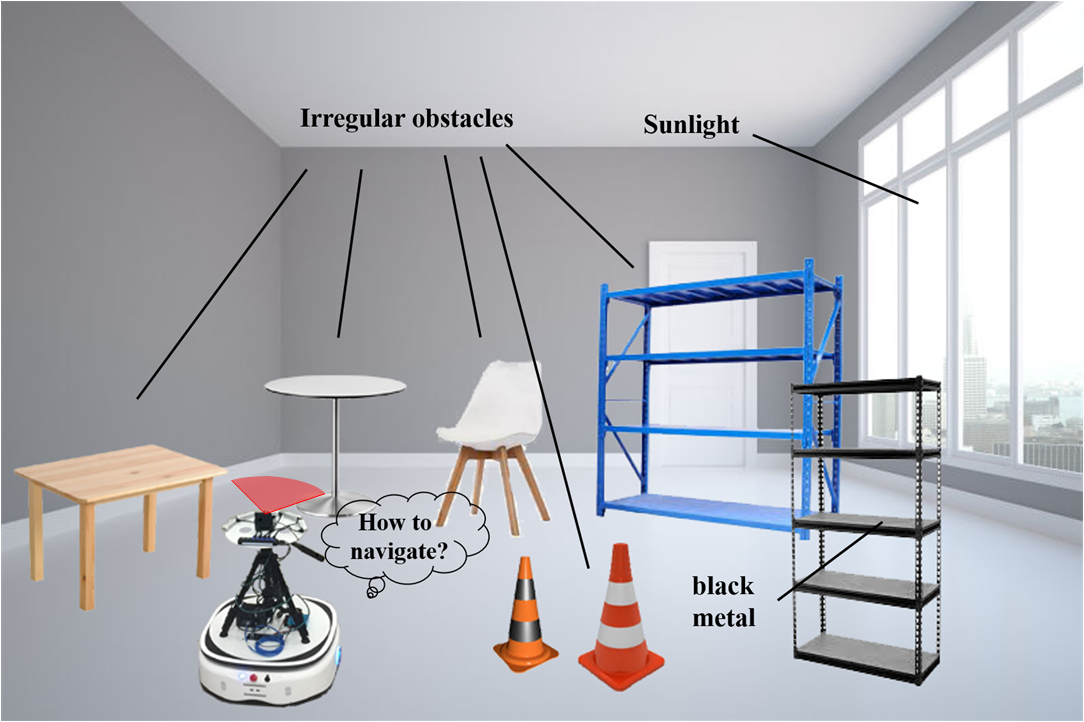}
	\vspace{-0.1cm}
	\caption{
		One-dimensional laser sensors cannot fully perceive this scene, including irregular obstacles \emph{e.g.}, tables, chairs, shelves, and non-reflectional material such as black metal.}
	\label{fig:teaser}
	\vspace{-0.7cm}
\end{figure}

The laser data possesses the advantage of accuracy and simplicity, while it contains less redundant information than image data, which benefits the training of reinforcement learning models. However, laser sensors notoriously have several shortcomings. First, one-dimensional laser observations cannot exploit the semantic information in the scene, and they are not able to access the shape of obstacles. Therefore, the one-dimensional laser device is difficult to model irregular objects in space (as shown in Fig.~\ref{fig:teaser}). Second, the non-reflection of special materials in the environment also degrades the robustness of the laser sensor. Besides, its high price makes it impossible to deploy in large-scale real-world applications.

Based on these problems, we propose a vision-based framework to handle collision avoidance during robot navigation, which only relies on a single RGB monocular camera.
Compared with the laser sensor and RGB-D camera, the RGB monocular camera has several advantages: lower cost and more robust under complex scenarios (\emph{e.g.}, strong sunlight or environment with non-reflectional material). More importantly, the RGB sensor can provide rich semantic information about the environment, which can be utilized to deal with irregular obstacles. But, as mentioned, the image data captured by RGB cameras contain redundant information, leading to a large sim-to-real gap.


To utilize the advantage of RGB cameras while reducing the sim-to-real gap, we propose a novel pseudo-laser data computed from the single RGB image captured by a deployed monocular camera to utilize the advantage of semantic information while reducing the sim-to-real gap. Unlike~\cite{wang2019pseudo} directly mapping the depth map to Pseudo-LiDAR to improve the accuracy of 3D bounding box prediction, we use semantic segmentation to encode the depth map to obtain the simple and accurate one-dimensional pseudo-laser data. It encodes the rich information of the complex objects and the scene, which thus can provide more reliable input for DRL than the laser-based measurement. In particular, our approach for obtaining pseudo-laser data is composed of two stages, semantic masking, and deep slicing.

Due to the perspective view of the camera, although the ground often takes up a large amount of space, it has no substantive significance for obstacle avoidance. Previous researchers~\cite{choi2019deep} usually project a depth map as a point cloud to remove useless ground information. However, the mapping method may fail in certain complex scenes, \emph{e.g.}, water on the ground, clothes on the floor, and slopes. Therefore, we exploit the semantic information of the scene to infer the traversable region, which considers each pixel information in the field of view (FOV). In this way, we design the semantic segmentation mask, which can remove the traversable region depth information from the estimated depth map. In addition, we can also extract the edge contours of some irregular objects to improve the perception of irregular objects through semantic information. In practice, the semantic segmentation mask can be simplified as a binary map indicating the pixels belonging to the traversable region or not.

After removing the traversable region from the depth map, we introduce a dynamic local minimum pooling operation to obtain our proposed pseudo-laser data. A naive way to generate pseudo-laser data is to slice the depth map along a horizontal axis. However, this slicing approach ignores the shape of obstacles. For instance, as shown in Fig.~\ref{fig:teaser}, if only slicing the feet of the table, the robot may collide with the upper part of it. In contrast, the dynamic local minimum pooling operation can extract the depth information of each dimension from the depth semantic map and fuse it into one-dimensional pseudo-laser data, in which each value is the nearest distance on the vertical axis to effectively avoid some irregular objects in scenes.

With the pseudo-laser data, we adopt a DRL framework to sample the collision avoidance strategy of the robot to efficiently steer away from obstacles. In particular, limited by the FOV of the camera, collision avoidance during navigation is formulated as a partially observable Markov decision process \cite{spaan2012partially}. To enhance the performance of the model, we incorporate an attention mechanism \cite{hu2018squeeze} to allow the agent to focus on the salient objects at the current moment. In this way, it can handle various types of objects (\emph{e.g.}, static objects and dynamic objects) differently during navigation. Besides, we introduce the LSTM \cite{hochreiter1997long} submodule to leverage the historical states of the agent to model its temporal actions.

At the training stage, we design several simulation environments with different levels of difficulty, to optimize the process of DRL algorithms. The algorithm starts training from simple scenes and tasks, then migrates the trained model to more complex scenes and tasks. 
In order to transfer our model from the simulation environment to the real world, during the training process, we augment the observation collected from the simulator by adding specifically-designed noises to the training data, so that it can better simulate the pseudo-laser data in real-world scenes. 

The main contributions of our work are as follows:
\begin{itemize}
	\item We present a vision-based collision avoidance framework based on deep reinforcement learning for robot navigation, which only relies on a single RGB monocular camera as the sensor. 
	
	\item To share the advantage of laser sensors and utilize the semantic information of the environment, we use the RGB data to encode one-dimensional pseudo-laser data for collision avoidance. Concurrently, we propose a new slicing approach for generating pseudo-laser data based on semantic depth maps, which greatly improves the collision avoidance ability of agents against irregular objects.
	
	
	\item To accommodate our vision-based model, we introduce a new data augmentation to enhance the robustness of the model and reduce the sim-to-real gap. Experiments show that the robot moves smoothly through the adaptive data augmentation.
	
\end{itemize}

\begin{figure*}[t]
	\centering
	
	\includegraphics [height=58.8mm,width=175mm ]{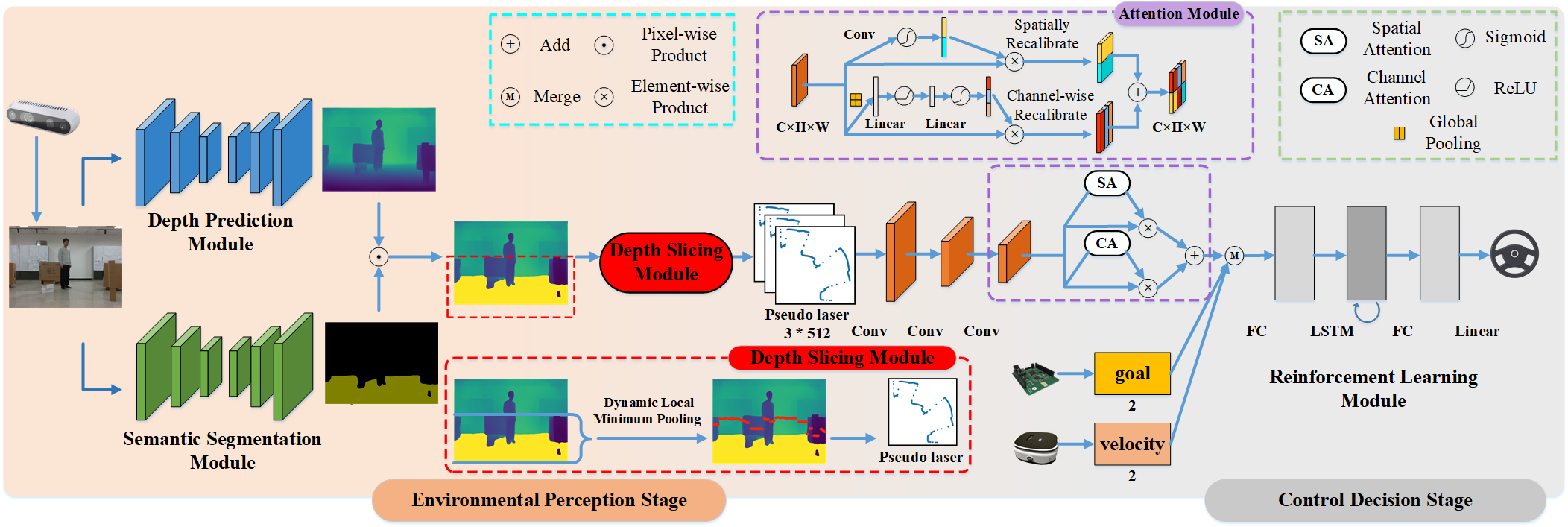}
	\vspace{-0.4cm}
	\caption{The pipeline of our entire framework. We fuse the depth and semantic information from the RGB image to generate pseudo-laser data, and send it to the DRL network we designed to output actions. }
	\label{fig:pipeline}
	\vspace{-0.7cm}
\end{figure*}

\vspace{-0.2cm}
\section{Related Work}
\vspace{-0.1cm}
\subsection{Laser-based navigation}

The artificial potential fields \cite{cosio2004autonomous}, dynamic window approaches \cite{fox1997dynamic}, these traditional navigation methods have achieved safe obstacle avoidance in simple scenarios. However, the map information needs to be known, and it takes a lot of time or even fails due to the sensitivity to hyperparameters and local minima in some complex dynamic scenes. For example, the advanced traditional method ORCA \cite{van2011reciprocal} needs to know the location information of other agents and static map information. Its generalization is poor in complex scenes due to tedious parameter adjustment.

In recent years, with the development of deep learning, \cite{michels2005high, fan2018crowdmove, chen2017socially} and \cite{fragkiadaki2015learning} have shown the effectiveness of learning-based navigation methods. \cite{pfeiffer2017perception} proposed a data-driven end-to-end motion planner. They used the data generated in the simulation environment through the ROS navigation package to train a model. It is able to learn the complex mapping from laser measurement and target positions to the required steering commands for the robot. This model can navigate the robot through a previously unseen environment and successfully react to sudden changes. Nonetheless, similar to the other supervised learning methods, the performance is seriously limited by the quality of the dataset. To solve the problem, \cite{tai2017virtual} proposed a mapless motion planner based on DRL to map 10-dimensional range findings to actions. The robot can safely navigate through a priori unknown environment without collision in this way. However, due to the sparse laser and the simple training scene, the model does not perform well on dynamic obstacles. Work \cite{long2018towards} proposed a multi-scenario and multi-stage training framework based on the PPO \cite{schulman2017proximal} reinforcement learning method. By inputting three consecutive frames from a 180-degree laser scanner, obstacle avoidance for dynamic objects is achieved, and it is extended to multi-robot applications. On this basis, \cite{choi2019deep} proposed an LSTM agent with Local-Map Critic (LSTM-LMC) to reduce the FOV to $90\,^{\circ}$ and achieve the performance of $180\,^{\circ}$. They used point cloud mapping and slicing to obtain laser data from the depth map. Due to the limitation of depth map slicing of point cloud mapping, the robot performs poorly on some complex ground (\emph{e.g.}, water surface, stairs, clothes, \emph{etc}.) and irregular obstacles. We continued the work of \cite{long2018towards} and reduced the FOV to $90\,^{\circ}$ in our way. And the robot has the ability to avoid obstacles to irregular objects through our pseudo-laser data.

\vspace{-0.2cm}	
\subsection{Vision-based navigation}
\vspace{-0.1cm}	
Due to laser scanners are expensive and can only capture limited information in the FOV. The camera can provide rich information about the operating environment of the robot and is low-cost, light-weight, and applicable for a wide range of platforms. There are a variety methods of vision-based navigation and obstacle avoidance, such as through optical flow \cite{souhila2007optical}, detection of vanishing points \cite{bills2011autonomous} and end-to-end mapping based on deep learning \cite{kim2015deep}. \cite{tai2016deep} proposed an end-to-end model that can directly map depth maps to actions. However, operating a mobile robot to collect training data in the real world is inconvenient and time-consuming. Works such as \cite{ma2019learning} and \cite{bharadhwaj2019data} trained policy for mapping from RGB image to action in the simulation environment. They perform well in training scenarios, but it is difficult to transfer to real-world environments. \cite{gordon2019splitnet} used Habitat scene renderer \cite{savva2019habitat} to train visual navigation models, thereby reducing the gap between sim-to-real. However, in unseen scenarios, the model needs to be retrained. \cite{chen2019learning} proposed a criterion for evaluating the gap between sim-to-real, and based on this proposed a visual obstacle avoidance method. However, due to the use of imitation learning and simple scenarios, the method is difficult to apply in challenging scenarios. Instead, we use a novel depth map slicing method that combines depth and semantic information to achieve mapping from a single RGB image to pseudo-laser data. The robot can safely avoid obstacles in more complex scenes in this way.

\vspace{-0.2cm}	
\section{Methodology}
\vspace{-0.1cm}


Since the DRL-based method can continuously interact with the environment in a `trial-error' manner, and can complete mapless navigation and obstacle avoidance, which has been proven in~\cite{long2018towards}, we choose DRL as our training framework. In order to adapt the agent to the environment with irregular obstacles, we generate pseudo-laser data from the RGB image as an input to the DRL to obtain more robust control actions. As shown in Fig.~\ref{fig:pipeline}, we obtain the corresponding depth map and semantic segmentation mask from the input RGB image. Then, we use the semantic segmentation mask to obtain a depth map culling out the traversable region from the image. After that, we propose a dynamic local minimum pooling operation that slices the depth map to obtain pseudo-laser data. Finally, we input the pseudo-laser data to the DRL network and it samples the actions to guide the agent to avoid obstacles. In the following, we first introduce the method we propose to generate pseudo-laser data and then our network architecture that integrates an LSTM and attention model.

\textbf{Deep reinforcement learning module.} In this part, we hope that agents can effectively avoid obstacles in complex unknown environments and rely on vision only to safely reach the location of a man-made designated target.

Affected by the FOV of the sensor, the agent cannot fully perceive the surrounding environment during the training process, thus we define the training process as a partially observable Markov decision process (POMDP). In simple terms, a POMDP is a cyclical process in which an agent takes action to change its state to obtain rewards and interact with the environment. POMDP consists of 6 tuples (\emph{S}, \emph{A}, \emph{P}, \emph{R}, \emph{$\Omega$}, \emph{O}) where \emph{S} is the state space, \emph{A} is the action space, \emph{P} is the transition probability, \emph{R} is the reward function, \emph{$\Omega$} is the observation space, and \emph{O} is the observation probability.

The goal of DRL is to learn the policy of the agent \emph{$\pi$}(\emph{a},\emph{o}) = \emph{p}(\emph{a}$\mid$\emph{o}) that maximizes discounted return
\vspace{-0.2cm}
\begin{equation}G = \sum_{t = 0}^\infty \gamma^t \mathbb{E}[r(s_t,a_t)] \vspace{-0.2cm} \end{equation}

where $\gamma$ is the discount factor for future rewards. Because the action space of robots is continuous, we adopt the Actor-Critic (AC) framework based on policy gradient to implement our DRL obstacle avoidance policy.
\begin{itemize} 	\item \textbf{Observation space.} 	For the observation $o$ of the agent, we use the pseudo-laser measurement with $90\,^{\circ}$ horizontal FOV.
	\item \textbf{Action space.} For the action $a$ of the agent, we use R$^{2}$ vectors for linear and angular velocities. Linear velocity of the agent is in range [0,1]m/s and angular velocity is in range $[-90,90]\,^{\circ}$/s. We use normalized velocities which are in range $[-1,1]$ as outputs of the neural networks.
	\item \textbf{Reward function.} Our objective is to avoid collisions during navigation and minimize the mean arrival time of robots. The reward function follows \cite{long2018towards}.
\end{itemize}  
\vspace{-0.1cm}
\begin{equation} 
r_{i}^{t} = (r_{goal})_{i}^{t}+(r_{collision})_{i}^{t}+(r_{rotational})_{i}^{t} 
\vspace{-0.1cm}
\end{equation}  
The total reward $r$ received by robot $i$ at timestep $t$ is a sum of three terms, $r_{goal}$, $r_{collision}$, $r_{rotational}$.
$r_{goal}$ represents the reward of whether the agent has reached its goal. If the agent collides, $r_{collision}$ is a penalty. And $r_{rotational}$ encourages the agent to move smoothly.
\vspace{-0.3cm}

\begin{small}
	
	\begin{equation}
	(r_{goal})_{i}^{t} = \left\{ \begin{array} { l l } { r _ { arrival } } & { \text { if } \| \textbf{p} _ { i } ^ { t } - \textbf{g} _ { i } \| < 0.1 } \\ { \omega _ { g } ( \| \textbf{p} _ { i } ^ { t-1 } - \textbf{g} _ { i } \|-\| \textbf{p} _ { i } ^ { t } - \textbf{g} _ { i } \| ) } & { \text { otherwise } } \end{array} \right.
	\end{equation}
	
\end{small}

${\textbf{p} _ { i } ^ { t }}$ is the position of robot $i$ at time $t$ and $\textbf{g} _ { i }$ is the goal position of robot$i$.
In order to avoid collisions between agent and obstacles, we set $r_{collision}$ as a penalty.

\vspace{-0.3cm}

\begin{equation}
(r_{collision})_{i}^{t} = \left\{ \begin{array} { l l } { r _ {  collision} } & { \text { if } \| \textbf{p} _ { i } ^ { t } - \textbf{p} _ { j } ^ { t } \| < 2 R } \\ { } & { \text { or } \| \textbf{p} _ { i } ^ { t } - \textbf{B} _ { k } \| < R } \\ { 0 } & { \text { otherwise } } \end{array} \right.
\end{equation}

$\textbf{B} _ { k }$ is the position of obstacle $k$ and $R$ is the radius of the robot. 
To encourage the robot to move smoothly, a small penalty $(r_{rotational})_{i}^{t}$ is introduced to punish the large rotational velocities.
\vspace{-0.15cm}
\begin{equation}
(r_{rotational})_{i}^{t} = \omega _ { w } | w _ { i } ^ { t } | \quad \text { if } | w _ { i } ^ { t } | > 0.7
\end{equation}

We set $r_{arrival}$ = 15, $\omega_{g}$ = 2.5, $r_{collision}$ = -15 and $\omega_{w}$ = -0.1 in the training procedure.

Our DRL framework is based on the module of \cite{long2018towards}. Compared with their method, the FOV of our framework is much smaller (\emph{i.e.}, $90\,^{\circ}$) and the observation may be noisier. In order to make up for the limitation of the sensor FOV, we introduce LSTM into the decision policy network. At the same time, we design a FEG module to weight the input pseudo-laser measurement to further improve the ability of robots to understand and capture the surrounding environment.


\textbf{Pseudo-laser data.} In this work, we attempt to deploy an RGB monocular camera to the agent to perform collision avoidance. Compared with other sensors, the RGB camera is low-cost and can provide rich semantic information in most scenes. However, compared with laser sensing data, it may contain redundant information. To share the advantage of the laser sensor and utilize the obtained semantic information, we propose a ``pseudo-laser data" generated from the RGB image as our input to the DRL network to output the actions. Our approach is inspired by the prior work~\cite{long2018towards} that uses the laser sensor to obtain the distance measurement away from the nearest obstacles and formulates them as a one-dimensional vector. The major advantage of this method is that they do not need to have the perfect sensing for the environment to perform collision avoidance, which significantly reduces the sim-to-real gap. The motivation of pseudo-laser data we proposed is to ignore the difference between simulation and reality. As simple and accurate one-dimensional data, pseudo-laser can also reduce the training difficulty of DRL while incorporating rich semantic information to improve robustness to irregular obstacles.

\textbf{Depth map.} To accomplish this, we use the depth estimation model FastDepth \cite{wofk2019fastdepth} and the image semantic segmentation model CCNet \cite{huang2019ccnet} as two parallel branches to obtain the depth map $M_D$ and semantic segmentation mask $M_G$ of the input RGB image $I$ ($I\in \mathbf{R}^{H\times W}$), respectively. For indoor and outdoor scenes, we train the depth estimation branch on the NYU \cite{silberman2012indoor} and KITTI \cite{geiger2013vision} datasets, respectively. Intuitively, since the traversable region information cannot assist agents to avoid obstacles, we need to get the semantic depth map. To achieve this purpose, we leverage the second branch to produce a semantic segmentation mask providing pixel-level traversable region labels. Since the only concern at this stage is ``which pixel in the image belongs to the traversable region", our second branch can provide a binary mask $M_G$, indicating which pixel of the image belongs to the traversable region, \emph{i.e.}, $M_G(i,j)=0$ if the pixel at $(i,j)$ belongs to the traversable region and vice versa. Finally, we obtain a depth map without traversable region depth via $M = M_D \odot M_G$, where $\odot$ refers to the pixel-wise multiplication.


\textbf{Depth map slicing.} In order to obtain the pseudo-laser data $L$, the naive solution is to slice the depth map along a specific row, \emph{i.e.}, $L = M(i,:)$, where $i$ indicates the index of the row and thus $L\in \mathbf{R}^{1\times W}$. However, this solution may not be robust for irregular obstacles. For instance, if only slicing the feet of the table, the robot may collide with the upper part of it. When there exists a small obstacle, such a naive slicing method may cause a similar problem. To address this problem, we propose a simple yet effective solution, \emph{i.e.}, dynamic local minimum pooling operation. In practice, we only keep the lower half of the depth map for computation, \emph{i.e.}, $\hat{M} = M(\frac{H}{2}:H, :)$, as shown in Fig.~\ref{fig:pipeline}. This is because the upper half of the image often provides little context information for collision avoidance. Then, we perform column-wise minimum pooling to select the minimal value along each column, \emph{i.e.}, the pseudo-laser data $L$ can be computed as $L(j)=\min \hat{M}(:,j), \forall j=\{1,\cdots, W\}$), where $\hat{M}(i,j) \ne 0$. Intuitively, the minimum pooling operation approximately localizes the nearest obstacles of any height in the environment. Hence, the extracted pseudo-laser data will be forwarded to the DRL network.


Our DRL module has the same reward function and action space as the work~\cite{long2018towards}. Compared with their method, the FOV of our framework is much smaller (\emph{i.e.} $90\,^{\circ}$) and the observation may be noisier. To address the limitations, we introduce the attention mechanism and LSTM into the decision policy network.

\textbf{Attention.}
In the DRL network, we take three consecutive pseudo-laser data as input, \emph{i.e.} $\{L_{t-2}, L_{t-1}, L_t\}$. Such input can enable our agents to infer the motion of surrounding objects, and thus make more effective collision avoidance behavior. Due to the limited FOV, the policy network needs to extract the observation more effectively. Besides that the convolutional layers extract visual features of the pseudo-laser data, we allow the feature to pass through a self-attention model~\cite{roy2018concurrent} to recalibrate the spatial and channel-wise features. The recalibrated spatial and channel-wise features will be merged as the observation feature $O$.


\textbf{LSTM.} As shown in Fig.~\ref{fig:pipeline}, we introduce the LSTM into our network to solve the limited FOV problem. LSTM has been applied as a temporal model to address the long-term dependency issue \cite{hochreiter1997long}.
Here, LSTM allows the agent to make more reasonable actions based on the current state and the historic states. In specific, we add LSTM behind the fully connected layer where the behavioral state (\emph{i.e.} the goal $\textbf{g}$ and velocity $\textbf{v}$ of the robot) and the observation $O$ are merged. The input of LSTM is a 256-dimensional feature and the output is also a 256-dimensional feature. Finally, the feature will be passed through another fully-connected layer and to map an action.

\textbf{Training.}
Our training is conducted in a virtual environment, in which agents interact with the environment and receive rewards from the environment to guide their learning. If the environment is too complex, or the task is too difficult, it will cause the DRL model to converge too slowly. In order to improve the efficiency of algorithm training, following the multi-stage multi-scenario training strategy~\cite{long2018towards}, we first train our DRL model in simple scenarios for simple tasks and then transfer the trained model to a more complex environment for further training.

\begin{figure}[t]

	\centering
	
	\includegraphics [height=37.21mm,width=75mm ]{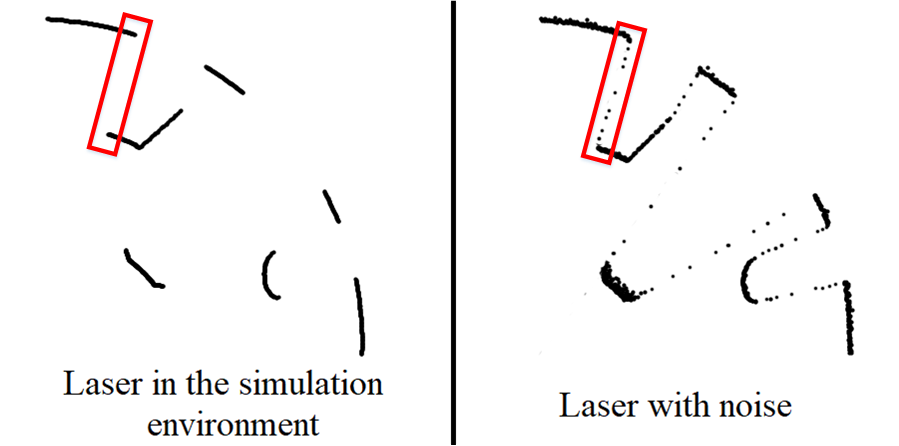}
	\vspace{-0.3cm}
	\caption{The raw laser measurement captured in the virtual scene and the noisy laser measurement after our data augmentation.}
	\label{fig:noise}
	\vspace{-0.7cm}
\end{figure}

\textbf{Data augmentation.}
To enhance the robustness of the DRL-based model in the real world, we add specific noise to the training laser measurement collected in the simulation environment to reduce the sim-to-real gap.  
In real-world scenes, based on observations, the measurement errors often occur around the boundaries of objects if some parts of one object block the other, \emph{i.e.}, the depth values of the boundary of the blocking part may be uncertain. Therefore, during training, we intend to add noise to emulate such measurement errors in order to produce adversarial training data.

To identify the junction boundary from the training laser measurement, \emph{i.e.}, a one-dimensional vector with precise depth information, we simply assume that, if the difference of two adjacent values in the vector is larger than the threshold $\alpha$, there may exist a junction boundary. To add noise, we locate the neighborhood of the junction boundary in the vector and replace all the values within the neighborhood by linearly interpolating the end points of the neighborhood (see Fig.~\ref{fig:noise}). 
On the other hand, for each vector element outside the neighborhood of the junction boundary, we simply add Gaussian white noise whose variance is proportional to the vector value.


\vspace{-0.3cm}
\section{Experimental Results}
\vspace{-0.12cm}
In this section, we first test the control decision stage and evaluate its performance. Then, we conduct experiments on the entire vision-based collision avoidance framework. Finally, we apply our model to a robot in real-world scenarios.

\begin{figure}[b]
	\centering
	\vspace{-0.6cm}
	\includegraphics [height=54.22mm,width=70mm ]{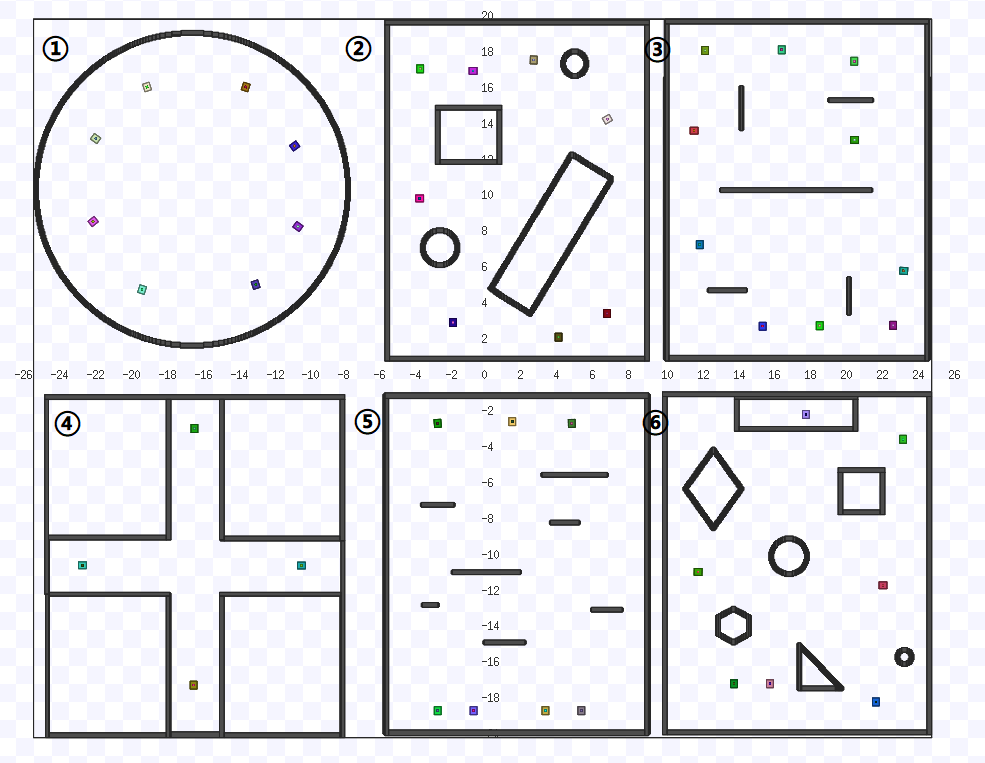}
	\vspace{-0.3cm}
	\caption{The training and testing scenarios built for the control decision stage in the Stage simulation environment. (1)-(3) are used for training. (4)-(6) are used for testing.}
	\label{fig:stage}
	\vspace{-0.3cm}
\end{figure}
\begin{figure}[b]
	\centering
	\vspace{-0.7cm}
	\includegraphics [height=25.34mm,width=85mm ]{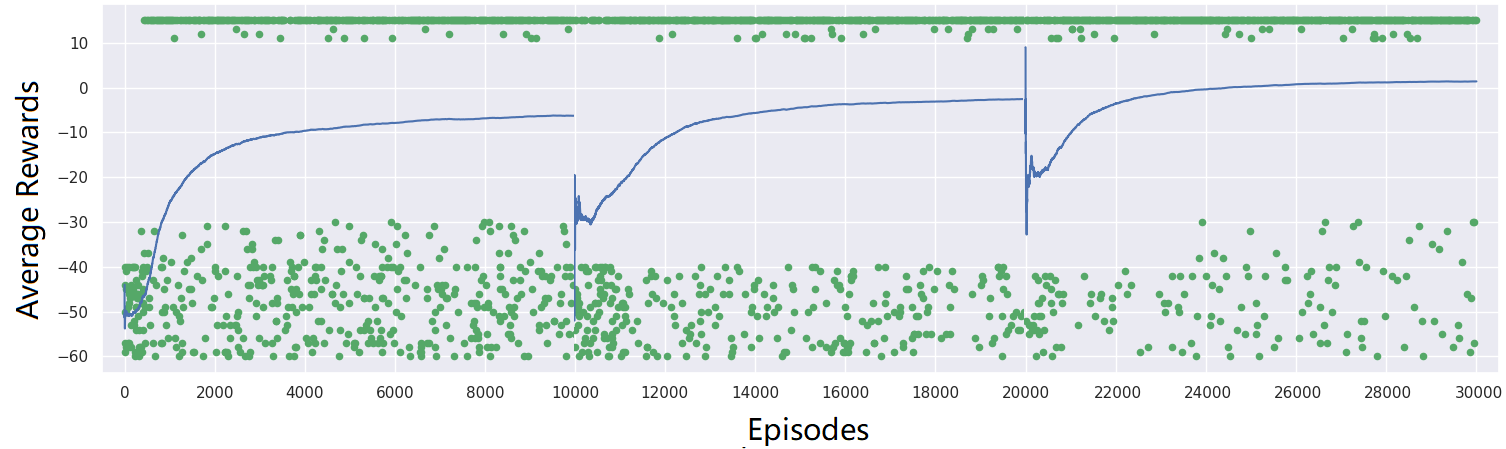}
	\vspace{-0.3cm}
	\caption{The average reward (blue curve) and individual rewards (green dots) during the three stages of training episodes.}
	\label{fig:convergence}
	
\end{figure}

\textbf{Implementation details.}
Our algorithm is implemented in Pytorch. We train the DRL policy on a computer equipped with an i7-7700 CPU and NVIDIA GTX 1080Ti GPU for approximately 40 hours. We build training and testing scenarios based on the Stage mobile robot simulator (as shown in Fig.~\ref{fig:stage}) and train 4 models with hyperparameters (see Table~\ref{para}) while adopting a multi-scene multi-stage alternating training method. The reward function follows \cite{long2018towards}. During training, the convergence of our reward function is shown in Fig.~\ref{fig:convergence}. Specifically, we train our model in three training scenes (see Fig.~\ref{fig:stage}) one after another. Note that we use a multi-agent collision avoidance task where each agent is assigned a shared collision avoidance policy, which can increase the robustness and training speed of our model. 

\begin{table}[t]	
	\setlength{\abovecaptionskip}{0.2cm}
	\setlength{\belowcaptionskip}{0.0cm}

	\caption{Hyper parameter settings in our experiments.}
	\label{para}
	\vspace{-0.5cm}
	\begin{center}
		\begin{tabular}{c c}
			
			\hline
			\textbf{Hyper parameter} & \textbf{Value} \\
			\hline
			Batch size & 1024 \\
			Maximum time steps & 150 \\
			Training episode & 2,000,000 \\
			Discount factor $\gamma$ & 0.99 \\
			Learning rate & 5e-5 \\ 
			LSTM unroll & 20 \\
			Target network update ratio & 0.01 \\
			KL penalty coefficient & 15e-4 \\
			the noise threshold $\alpha$ & 0.5 \\
			\hline
		\end{tabular}
	\end{center}
	\vspace{-0.8cm}
\end{table}

We use the following metrics to evaluate the performance of different navigation algorithms:

\begin{itemize}
	
	\item \textbf{Success Rate} - The number of times a robot reaching the target waypoint without collision and overtime.
	
	\item \textbf{Average Time} - The average time it takes for a robot to reach the goal safely. 
	
\end{itemize}

\begin{figure}[t]

	\centering
	
	\includegraphics [height=50.85mm,width=70mm ]{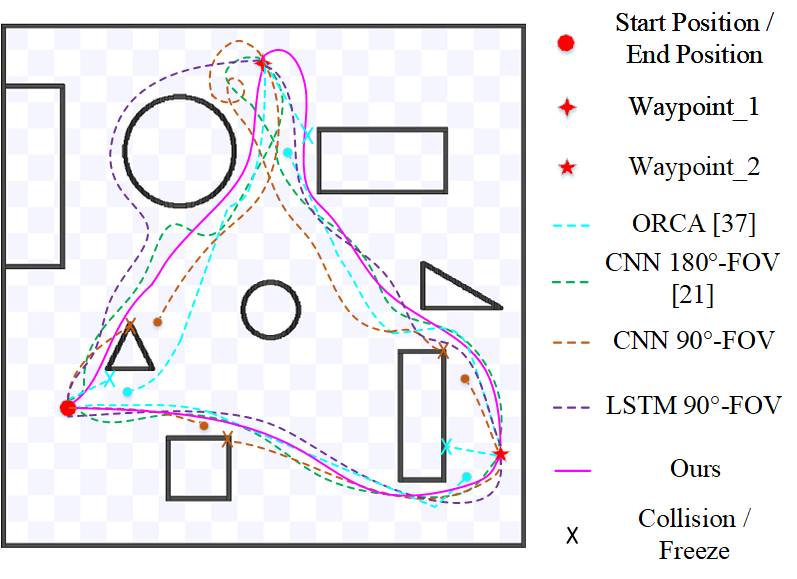}
	\vspace{-0.5cm}
	\caption{The trajectories generated by different models during navigating an agent in a virtual scene with two waypoints.}
	\label{fig:traj}
	\vspace{-0.4cm}
\end{figure}

\textbf{Network structure.} We evaluate 4 different network structures (CNN models with $180\,^{\circ}$ and $90\,^{\circ}$ FOV, LSTM based model and LSTM+Attention model with $90\,^{\circ}$-FOV) in DRL models and ORCA \cite{van2011reciprocal} as baseline in the test scenarios shown in Fig.~\ref{fig:stage}. Note that the CNN model with $180\,^{\circ}$-FOV is \cite{long2018towards}, while the CNN model with $90\,^{\circ}$-FOV is the same model yet trained with sensor observation of $90\,^{\circ}$-FOV.
The test scenario includes the collision avoidance tasks against static/dynamic obstacles: crossing, through walls, through fixed obstacles, and multi-robot obstacle avoidance, which are challenging scenes for evaluating the collision avoidance ability of our model against different types of obstacles. We randomly assign an initial position and a goal position to each robot and test it 100 times in different scenarios. The results are shown in Table~\ref{example_stage}. We also visualize the example trajectories of each model, as shown in Fig.~\ref{fig:traj}. Compared with peer models, our model can generate a smooth trajectory that successfully passes through two waypoints and reaches the goal without getting stuck or collisions.

We observe that our method is superior to the collision avoidance method ORCA and comparable to the model with $180\,^{\circ}$-FOV, while our model has a better success rate. For the $90\,^{\circ}$-FOV model, the success rate drops significantly. With the LSTM module introduced into the model, the performance of the model is improved. Based on this, we introduce channel attention and spatial attention, so that CNN can extract attention that is helpful for obstacle avoidance, which can also exceed the performance of the model with $180\,^{\circ}$-FOV. However, due to the complexity of the network structure, the robot may decelerate significantly in dangerous situations, resulting in an increase of average time.

\newcommand{\tabincell}[2]{\begin{tabular}{@{}#1@{}}#2\end{tabular}}

\begin{table}[t]

	\scriptsize	
	\setlength{\abovecaptionskip}{0.2cm}
	\setlength{\belowcaptionskip}{0.0cm}
	
	\caption{\fontsize{6.8pt}{\baselineskip}\selectfont PERFORMANCE OF AGENTS WITH VARIOUS FOV AND ARCHITECTURES}
	\label{example_stage}
	\vspace{-0.5cm}
	\begin{center}
		\begin{tabular}{|c|c||c|c|c|c|c|}
			
			\hline
			\multicolumn{2}{|c||}{Model} & \tabincell{c}{ORCA \\ \cite{van2011reciprocal}} & \tabincell{c}{CNN \\ \cite{long2018towards}} & CNN & LSTM & \tabincell{c}{LSTM \\ + Attention \\ (Ours)}\\
			\hline
			\multicolumn{2}{|c||}{Field of View (FOV)} & - & $180\,^{\circ}$ & $90\,^{\circ}$ & $90\,^{\circ}$ & $90\,^{\circ}$\\
			\hline
			\multirow{2}{*}{Scene 4} 
			& \tabincell{c}{Success \\ rate} & 92\% & 98\% & 79\% & 87\% & \textbf{100\%} \\
			\cline{2-7}
			& \tabincell{c}{Average \\ time} & 7.516 & \textbf{6.331} & 8.035 & 6.589 & 6.785 \\
			\hline
			\multirow{2}{*}{Scene 5} 
			& \tabincell{c}{Success \\ rate} & 72\% & 86\% & 16\% & 78\% & \textbf{87\%} \\
			\cline{2-7}
			& \tabincell{c}{Average \\ time} & 16.844 & \textbf{10.882} & 23.549 & 11.752 & 12.135 \\
			\hline
			\multirow{2}{*}{Scene 6} 
			& \tabincell{c}{Success \\ rate} & 78\% & \textbf{95\%} & 63\% & 80\% & 94\% \\
			\cline{2-7}
			& \tabincell{c}{Average \\ time} & 18.551 & \textbf{13.563} & 25.662 & 14.216 & 14.058 \\
			\hline
		\end{tabular}
	\end{center}
	\vspace{-0.8cm}
\end{table}

\begin{figure}[t]
	\centering
	
	\includegraphics [height=55mm,width=67.46mm ]{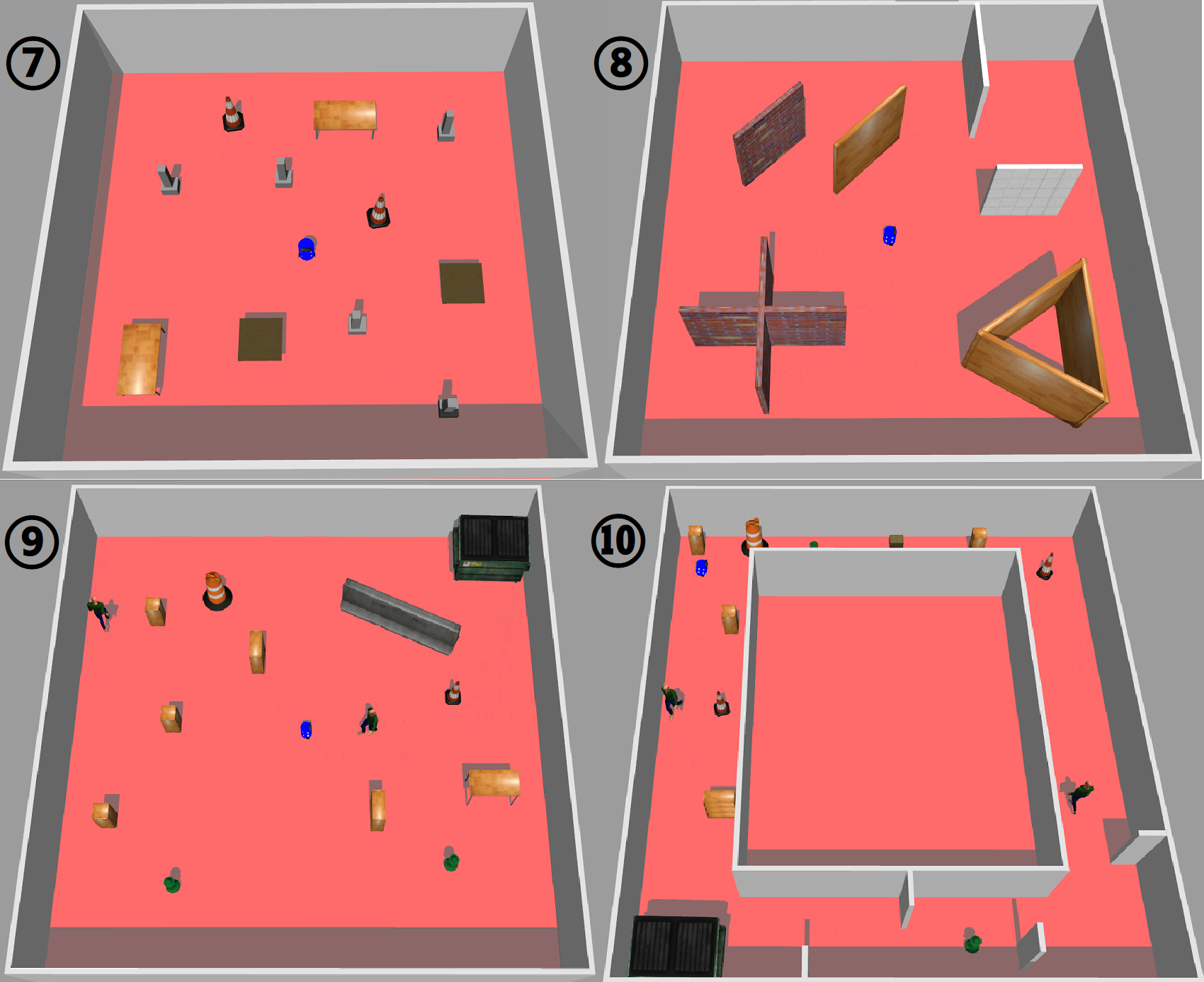}
	\vspace{-0.4cm}
	\caption{Different unseen 3D test scenarios are used for our experiments. (7) A scene consisting of only irregular objects. (8) A simple static scene. (9) A complex scene with walking persons. (10) A complex corridor scene.}
	\label{fig:gazebo}
	\vspace{-0.8cm}
\end{figure}


\textbf{The entire vision-based framework.} To demonstrate the effectiveness of our depth slicing method, we perform a visual simulation experiment. We build a 3D simulation environment based on Gazebo, as shown in Fig.\ref{fig:gazebo}. Scene 7 consists only of challenging irregular objects, while Scene 8-10 are complex corridor scenes with pedestrians. We build a simple two-wheel differential robot in Gazebo, deploying laser sensors and RGB-D cameras. The height of the robot is higher than the height of the table in the scene. In order to make the depth estimation and semantic segmentation models suitable for the Gazebo simulation environment, we simply collect scene-specific image data and pre-train the model for 6 hours so that it can be applied to virtual scenes. In order to improve the efficiency of image processing, we reduce the depth map by 5 times. The model runs on a computer with NVIDIA GTX 1080Ti GPU and an i7-7700 CPU, which can reach 10 frames per second.

\begin{table*}[t]
	\setlength{\abovecaptionskip}{0.2cm}
	\setlength{\belowcaptionskip}{0.0cm}
	
	\caption{\scriptsize PERFORMANCE OF AGENTS WITH DIFFERENT MODELS IN 3D SCENES OF DIFFERENT COMPLEXITY.}
	\label{example_gazebo}
	\vspace{-0.4cm}
	\begin{center}
		\begin{tabular}{|c|c|c|c|c|c|c|c|c|}
			
			\hline
			\multicolumn{2}{|c|}{Model} & \emph{E}$_{Btm.Las.}$ & \emph{E}$_{Top.Las.}$ & \emph{E}$_{Dep.1-D.}$ & \emph{E}$_{Dep.Pol.}$ & \emph{E}$_{Dep.1-D.Sem.}$ & \tabincell{c}{\emph{E}$_{Dep.Pol.Sem.}$ \\ (proposed)} & \tabincell{c}{\emph{E}$_{Dep.Sem.Noi.}$  \\ (proposed)} \\
			\hline
			\multirow{2}{*}{Scene 7} 
			& Success rate & 21\% & 15\% & 0\% & 35\% & 6\%  & 81\% & \textbf{89\%}\\
			\cline{2-9}
			& Average time & 15.856 & 17.536 & - & 13.612 & - & 13.251 & \textbf{12.898} \\
			\hline
			\multirow{2}{*}{Scene 8} 
			& Success rate & 100\% & 87\% & 76\% & 92\% & 87\% & \textbf{100\%} & \textbf{100\%}\\
			\cline{2-9}
			& Average time & \textbf{17.862} & 18.355 & 18.032 & 20.153 & 18.947 & 18.761 & 17.930 \\
			\hline
			\multirow{2}{*}{Scene 9} 
			& Success rate & 95\% & 87\% & 73\% & 93\% & 86\% & 93\% & \textbf{96\%} \\
			\cline{2-9}
			& Average time & 23.556 & \textbf{22.457} & 27.986 & 26.783 & 24.162 & 26.215 & 24.131 \\
			\hline
			\multirow{2}{*}{Scene 10} 
			& Success rate & 80\% & 75\% & 60\% & 75\% & 65\% & 80\% & \textbf{85\%} \\
			\cline{2-9}
			& Average time & \textbf{140.519} & 156.543 & 202.262 & 177.513 & 198.645 & 181.732 & 162.141 \\
			\hline
			
		\end{tabular}
	\vspace{-0.7cm}
	\end{center}
	
\end{table*}

We compare 7 models, lasers on the bottom of the robot (\emph{E}$_{Btm.Las.}$), laser on top of the robot (\emph{E}$_{Top.Las.}$), one-dimensional slice of depth map (\emph{E}$_{Dep.1-D.}$), dynamic local minimum pooling of depth map (\emph{E}$_{Dep.Pol.}$), one-dimensional slice of depth map combined with semantic mask (\emph{E}$_{Dep.1-D.Sem.}$), dynamic local minimum pooling of depth maps combined with semantic mask (\emph{E}$_{Dep.Pol.Sem.}$), and dynamic local minimum pooling of depth map and semantic mask with noise (\emph{E}$_{Dep.Sem.Noi.}$). We randomly generate a starting position and a goal position for a robot in each scene and test it 100 times. The corridor scene is set with 4 waypoints, and the robot moves circularly in the scene and tests for 20 times. The results are shown in Table~\ref{example_gazebo}.

We observe that, in Scene 7 consisting only of irregular objects, almost all the comparison models fail to accomplish the task. Only our slicing method takes the shape of the obstacle into account and achieves the best performance. Since the model \emph{E}$_{Btm.Las.}$ considers the shape of the bottom of obstacles only, it reacts poorly towards obstacles such as desks. Similarly, the model \emph{E}$_{Top.Las.}$ can only scan obstacle information at a certain height, which is not able to handle lower objects and cone objects. The model \emph{E}$_{Dep.1-D.}$ does not perform well for all objects since it falsely integrates the depth of the traversable region. The model \emph{E}$_{Dep.Pol.}$ shows some successful collision avoidance behaviors for some objects. However, the traversable region information is not considered, so the navigation success rate is still poor. Although the model \emph{E}$_{Dep.1-D.Sem.}$ considers the depth of the traversable region, it still only captures one-dimensional data, which is not effective. In contrast, the model \emph{E}$_{Dep.Pol.Sem.}$ considers both the depth of traversable region and the overall shape of the obstacle, which can achieve a better success rate of obstacle avoidance. We also test in more complex scenes, \emph{i.e.}, Scene 8-10, and the results show that our method has better robustness, and even better than laser sensors. Based on this, we also introduce a model trained via our proposed data augmentation method \emph{E}$_{Dep.Sem.Noi.}$. During training, the data augmentation setting details are as below. For identifying junction boundaries in laser measurements, we set the threshold for adjacent values as 0.5, and the range of neighborhoods is set as 8. For non-boundaries, we add the Gaussian white noise to each value of the laser measurement for data augmentation. The scale of the noise is 0.07 times the value. Table~\ref{example_gazebo} shows that the addition of noise in data augmentation for training the model has the highest success rate. Without such data augmentation, the agent will behave hesitantly and it may rotate at a large angle due to the difference between the pseudo-laser estimated from the depth map and the accurate laser sensing data in the simulator.

\begin{figure}[t]
	\setlength{\tabcolsep}{4pt}\small{
		\begin{tabular}{ccc}
			
			(A) & (B) & (C) \\
			\textbf{Scene with water} & \textbf{Scene with slope} & \textbf{Scene with clothes}  \\

			\includegraphics[height=18.0mm,width=25mm]{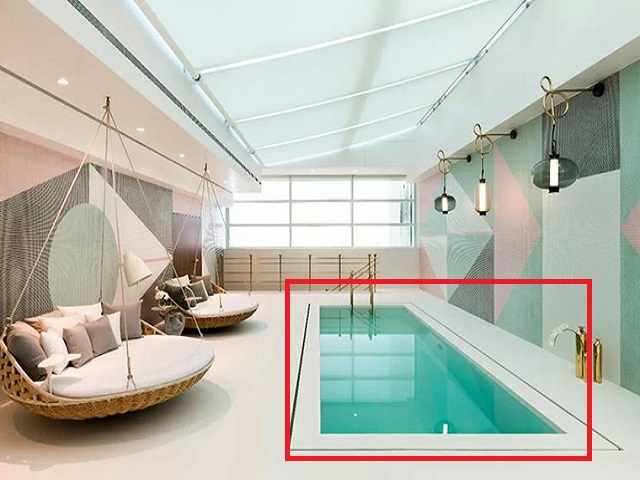}  &
			\includegraphics[height=18.0mm,width=25mm]{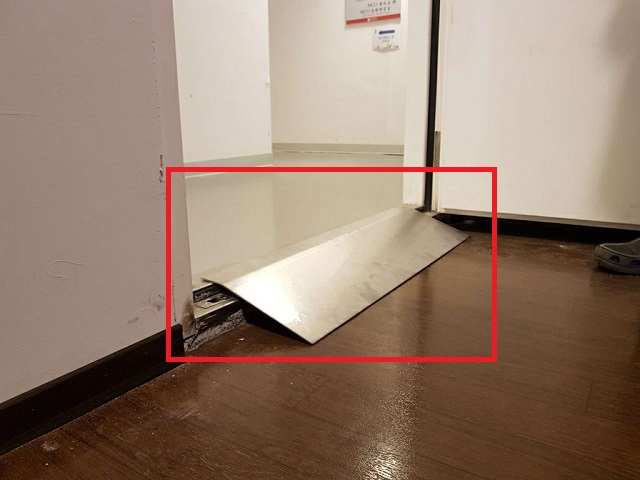}  &
			\includegraphics[height=18.0mm,width=25mm]{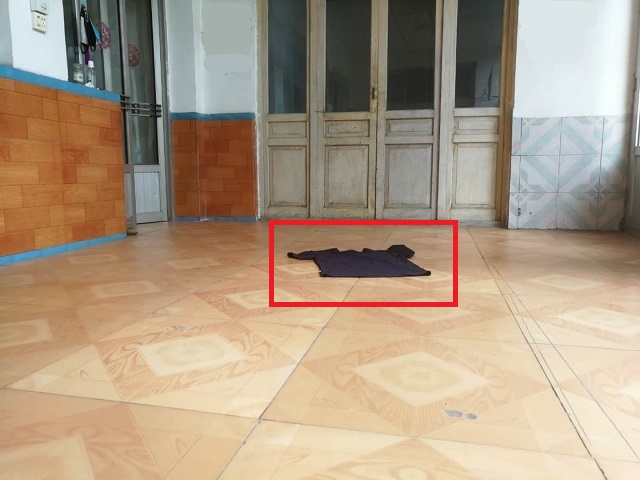} \\
			
			RGB image & RGB image & RGB image  \\
			
			\includegraphics[height=18.0mm,width=25mm]{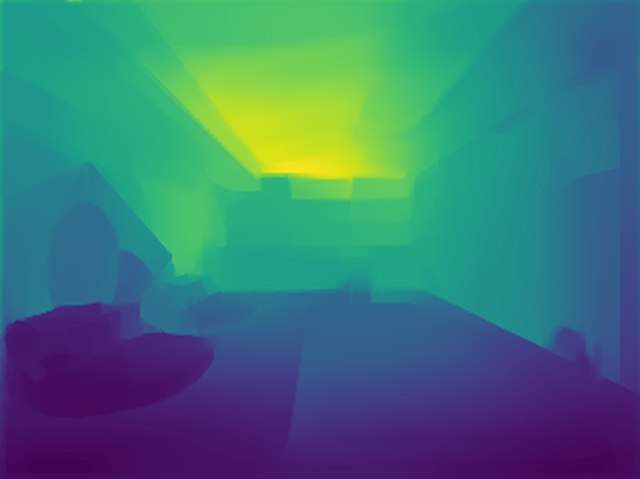}  &
			\includegraphics[height=18.0mm,width=25mm]{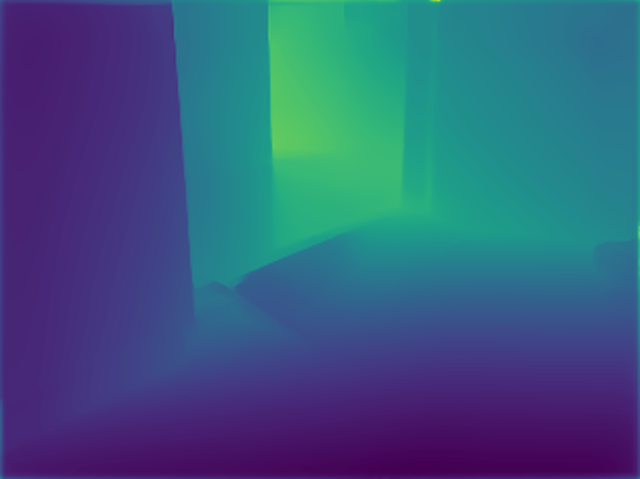}  &
			\includegraphics[height=18.0mm,width=25mm]{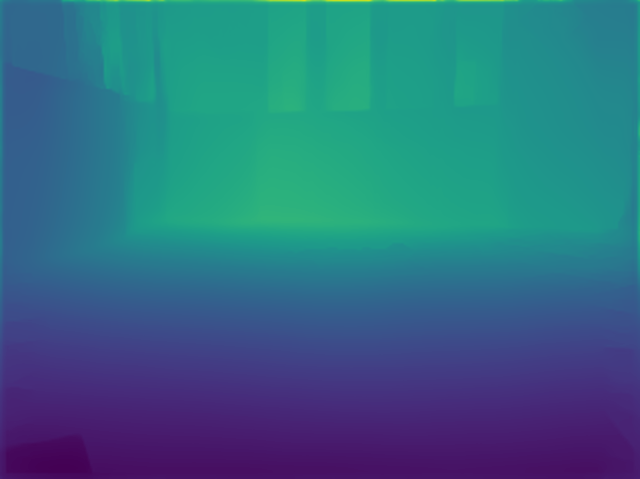} \\
			
			Depth image & Depth image & Depth image  \\

			\includegraphics[height=18.0mm,width=25mm]{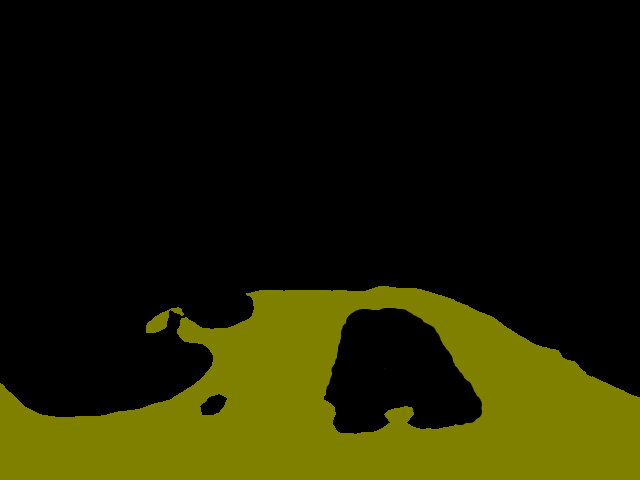}  &
			\includegraphics[height=18.0mm,width=25mm]{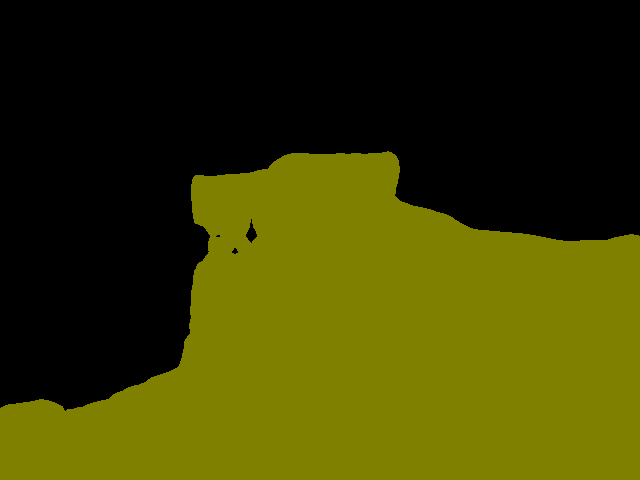}  &
			\includegraphics[height=18.0mm,width=25mm]{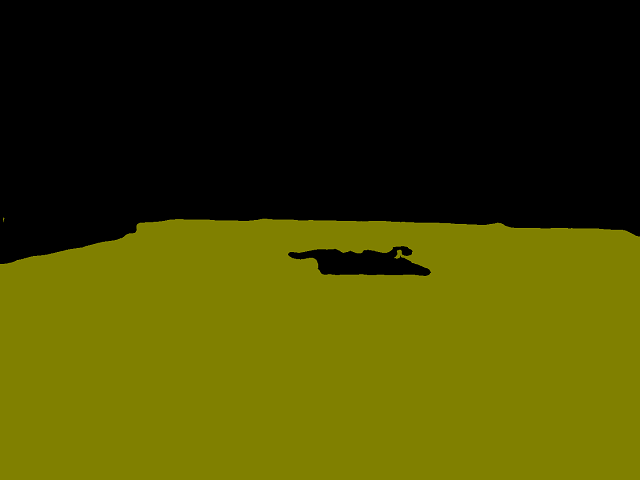} \\
			
			Semantic image & Semantic image & Semantic image  \\
			
			\includegraphics[height=18.0mm,width=25mm]{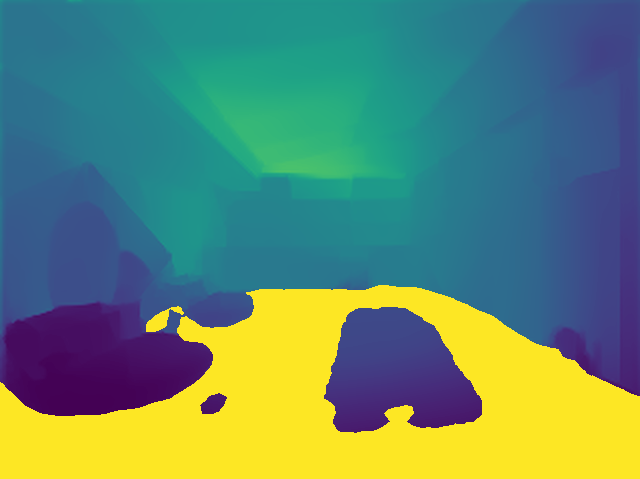}  &
			\includegraphics[height=18.0mm,width=25mm]{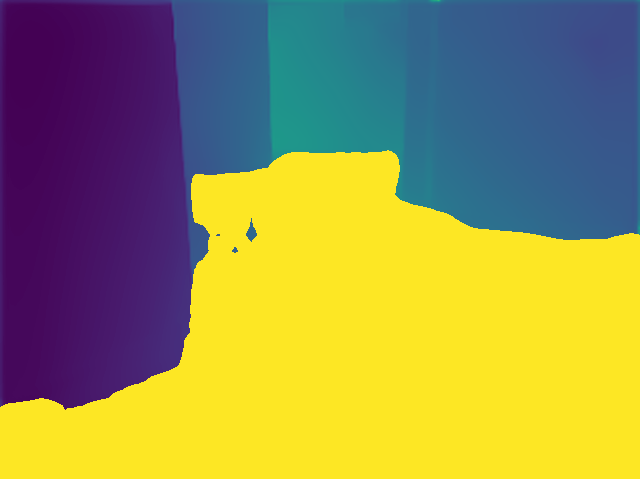}  &
			\includegraphics[height=18.0mm,width=25mm]{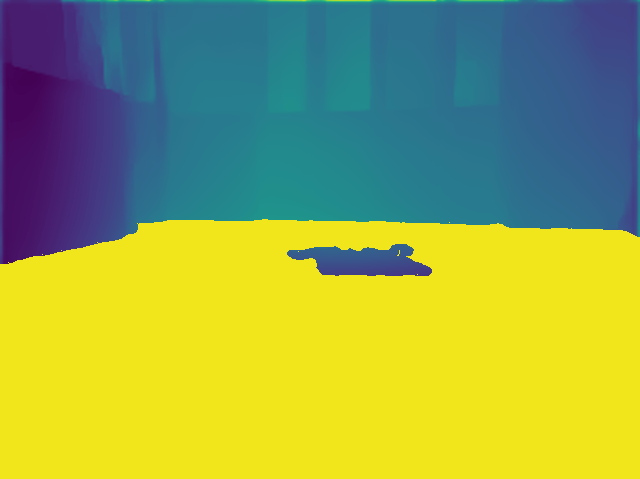} \\
			
			\footnotesize Semantic depth image & \footnotesize Semantic depth image & \footnotesize Semantic depth image  \\
			
			\includegraphics[height=18.0mm,width=25mm]{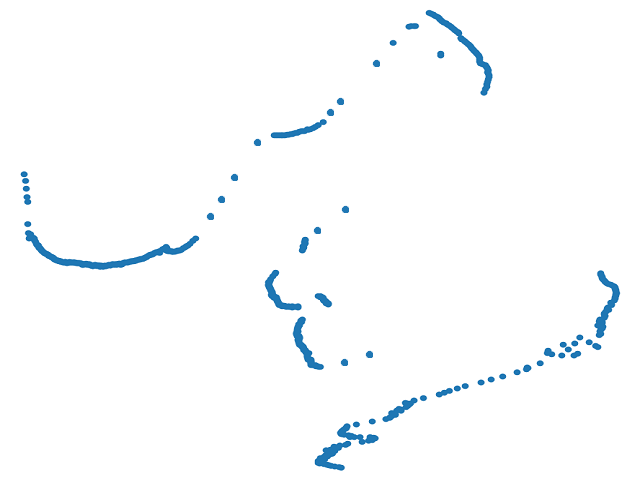}  &
			\includegraphics[height=18.0mm,width=25mm]{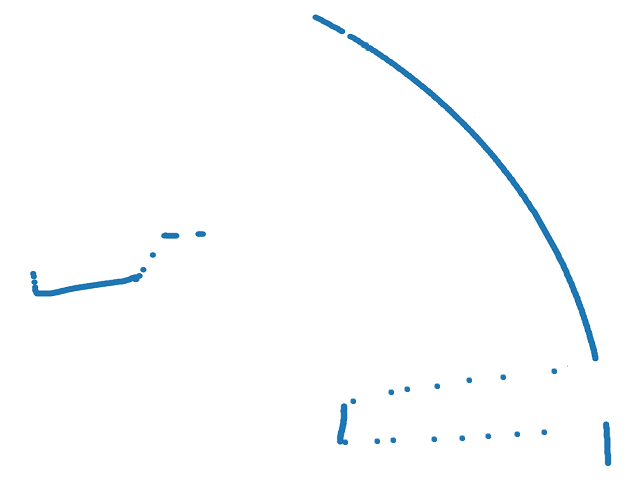}  &
			\includegraphics[height=18.0mm,width=25mm]{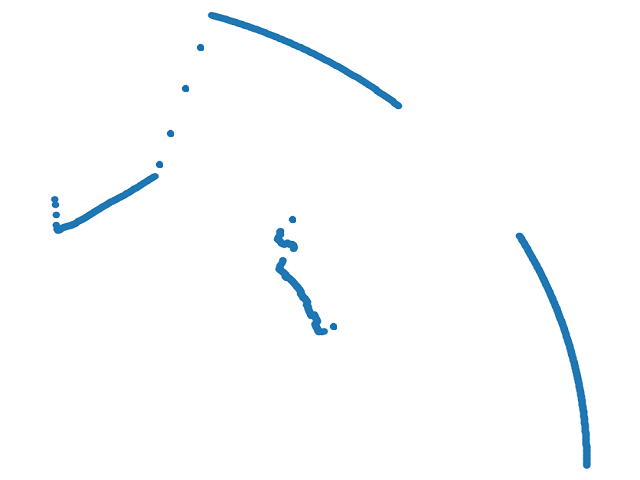} \\
			
			Pseudo-laser data & Pseudo-laser data & Pseudo-laser data  \\
			
	\end{tabular}}
	\vspace{-0.2cm}
	\caption{Visualization of the process of acquiring pseudo-laser data using semantic deep slicing in some complex scenarios. }
	
	\label{fig:semantic}
	\vspace{-0.7cm}
\end{figure}

\vspace{-0.1cm}
\textbf{Semantic information analysis.}
In order to further explain the role of our fusion of semantic information, as shown in Fig.~\ref{fig:semantic}, we visualize the process of acquiring pseudo-laser data in three complex scenarios. Taking Fig.~\ref{fig:semantic}(A) as an example, we obtain a depth map from an RGB image and a binary classification semantic mask with only the traversable region and background. After fusing into a semantic depth map, we use the depth slicing method we proposed to obtain the final pseudo-laser data. The traditional method obtains laser data from the depth map by horizontal slice mapping of the point cloud. Since semantic information is not considered, the water surface is also cut off from the ground. However, in reality, the water surface is not the safe passable region of the robot. Therefore we use semantic information to view the water as an unpassable region. The depth information of the water surface is also taken into account during the slicing process, thus providing the robot with the safest pseudo-laser data. In Fig.~\ref{fig:semantic}(B), there is a highly misaligned slope. For a simple horizontal slice, it is difficult to set a threshold, resulting in ground depth affecting navigation. Through semantic information, we can obtain the robot's passable area, so as to obtain the most efficient and reasonable pseudo-laser data. There is a piece of clothing on the ground in Fig.~\ref{fig:semantic}(C). In fact, the clothing cannot be crushed and should be regarded as an obstacle. We can obtain the pseudo-laser data considering the depth of the clothing. Therefore, the method of obtaining the semantic mask of the traversable region through the semantic information and then performing depth slicing can enable the robot to effectively deal with scenes with the complex ground.

\vspace{-0.2cm}

\subsection{Hardware experiment}

\vspace{-0.1cm}

\begin{figure}[t]
	\centering
	
	\includegraphics [height=32.54mm,width=88mm ]{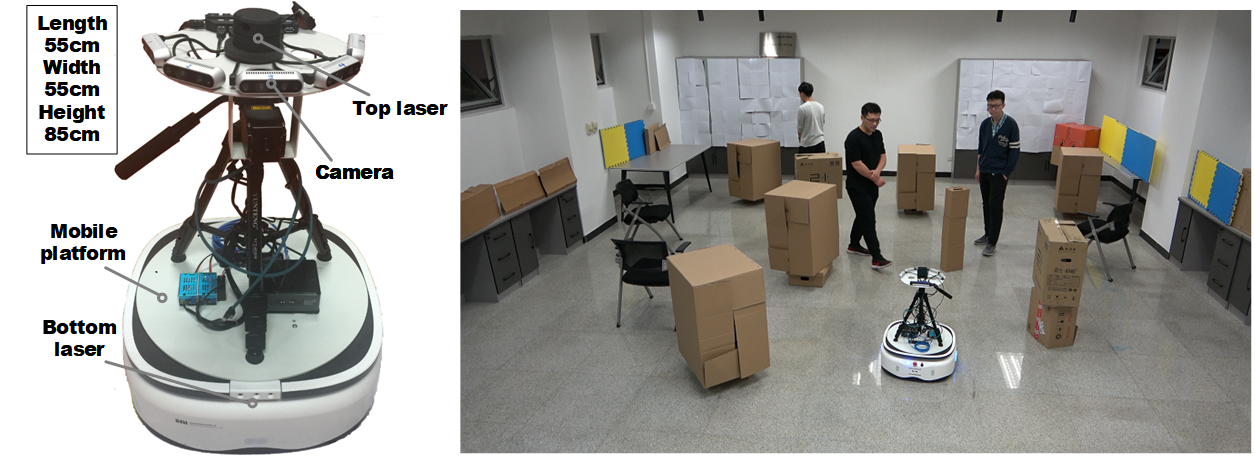}
	\caption{Illustration of the mobile robot platform used in our hardware experiments. It has 5 Intel Realsense D435 depth cameras, but we only use RGB images from \textbf{the middle one} of them. Two lasers at the top and bottom are used for testing.}
	\label{fig:robot}
	\vspace{-0.8cm}
\end{figure}

In order to see the overall performance of our method in the real world, we deploy a ROS-based mobile robot as shown in Fig.~\ref{fig:robot} in a real scene constructed of cardboard boxes, tables, and chairs. The robot transmits the acquired images through a router to a local host with an Nvidia GTX 1080Ti GPU. Then the host performs image processing action prediction locally and sends speed information to the robot. The image processing can reach 10 frames per second in real-time. Our method can safely navigate and avoid obstacles in unseen scenes, and shows good robustness and generalization to static irregular objects and challenging dynamic pedestrians. We have also proven the effectiveness of the noise module in real scenarios. See the supplementary video for more details.

\vspace{-0.2cm}

\section{Conclusion and Future work}
\vspace{-0.15cm}	
In this paper, we propose a framework that uses a low-cost monocular RGB camera to accomplish obstacle avoidance for mobile robots. Against the irregular obstacles in challenging environments, we design the pseudo-laser, which fuses distance and semantic information and exhibits good perception performance. We use data augmentation to improve the performance of the policy in the real world. We introduce attention and the LSTM module to solve the problem with limited FOV. However, our method has limitations. Due to the poor performance of the depth estimation method for the mirror, our method is invalid in some indoor scenes with mirrors. In future work, we will consider more complex concepts, such as robot positioning via pure vision \cite{hirose2019deep}.
	
	\vspace{-0.2cm}
	\addtolength{\textheight}{0cm}   
	

	


	

\end{document}